\pdfoutput=1

\documentclass[11pt]{article}
\usepackage{placeins}
\usepackage{acl2023}

\usepackage{times}
\usepackage{verbatim}
\usepackage{latexsym}
\usepackage{booktabs}
\usepackage{hyperref}
\usepackage{url}
\usepackage{subcaption}
\usepackage{graphicx}
\usepackage{etoolbox}
\usepackage{multicol}
\usepackage{amsmath}
\usepackage{amssymb}
\usepackage{mathtools}
\usepackage{amsthm}
\usepackage[T1]{fontenc}

\usepackage[utf8]{inputenc}

\usepackage{microtype}

\usepackage{inconsolata}

\newcommand{\txt}{\operatorname{text}}
\newcommand{\sound}{\operatorname{sound}}
\newcommand{\simm}{\operatorname{sim}}

\title{What Do Language Models Hear? \\ Probing for Auditory Representations in Language Models}

\author{
Jerry Ngo ~\;~ \hspace{1cm}
Yoon Kim~\;~
\vspace{1mm} \\
Massachusetts Institute of Technology 
\\
\normalsize 
\href{mailto:ngop@mit.edu}{\texttt{ngop@mit.edu}}~\;~ 
 \href{mailto:yoonkim@mit.edu}{\texttt{yoonkim@mit.edu}} 
\\
  }

\begin{document}
\maketitle
\begin{abstract}
This work explores whether language models encode meaningfully grounded representations of sounds of objects. We learn a linear probe that retrieves the correct text representation of an object given a snippet of audio related to that object, where the sound representation is given by a pretrained audio model.  This probe is trained via a contrastive loss that pushes the language representations and sound representations of an object to be close to one another. After training, the probe is tested on its ability to generalize to objects that were not seen during training. Across different language models and audio models, we find that the probe generalization is above chance in many cases, indicating that despite being trained only on raw text, language models encode grounded knowledge of sounds for some objects.

\end{abstract}

\section{Introduction}

\begin{figure}[h!]
\centering
\begin{subfigure}{0.45\textwidth}
    \centering
    \includegraphics[scale=0.24]{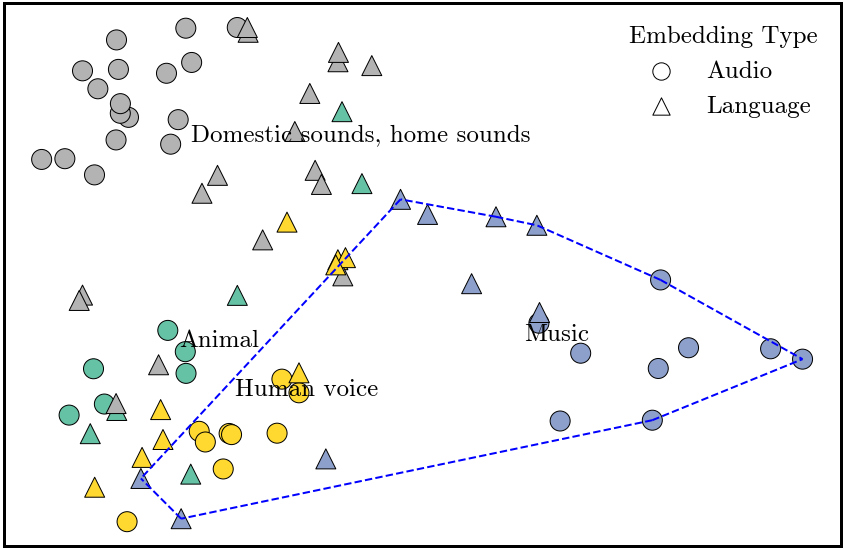}
\end{subfigure}
\begin{subfigure}{0.45\textwidth}
    \centering
    \includegraphics[ scale=0.24]{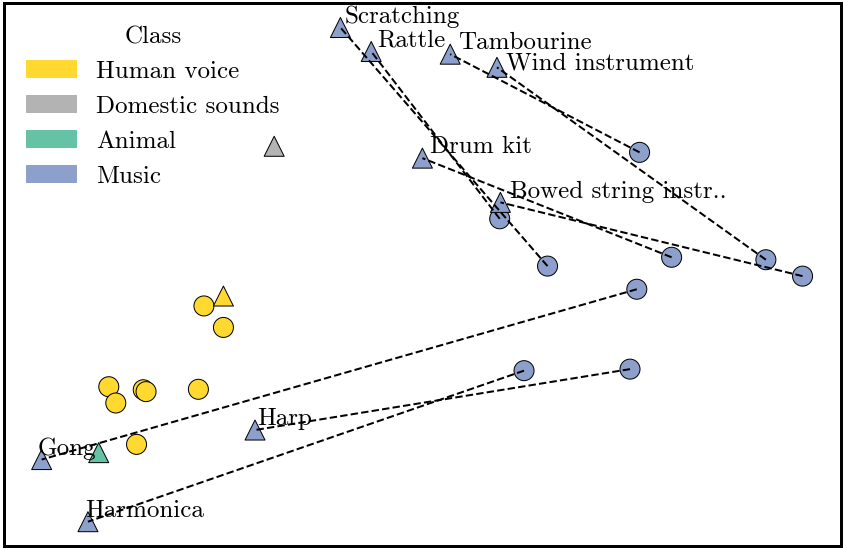}
\end{subfigure}
    \caption{(Top) Language (triangle) and sound (circle) representations aligned via Procrustes analysis \cite{schonemann_generalized_1966}, visualized via PCA. The language representation is from BERT \cite{devlin-etal-2019-bert} and the audio representation is from PaSST \cite{passt}. The classes are color-coded based on their parent nodes (i.e., \texttt{human voice}, \texttt{domestic sounds}, \texttt{animal}, \texttt{music}) according to the ontology from the FSD50K \cite{fonseca2021fsd50k}. (Bottom) A zoomed-in portion of the blue region of the top figure, which shows the structural similarities between the language and sound representations for the \texttt{music} category.}
       \label{fig:prelim}
 
\end{figure}

Despite being trained only on surface-form strings (i.e., without explicit grounding), language models (LMs) have been shown to learn representations of perceptual concepts that plausibly mirror the grounded, physical representations  of those same concepts. Examples of such concepts that have been investigated so far in the literature include color \cite{abdou-etal-2021-language}, direction \cite{patel2021mapping}, size \cite{zhang-etal-2020-language-embeddings,grand2022semantic}, geography \cite{konkol-etal-2017-geographical,lietard2021language,faisal-anastasopoulos-2023-geographic,chen2023more}, time \cite{gurnee2023language}, and even visual representations \cite{ilharco_probing_2021,merullo2022linearly,li_implications_2023}.  
The alignment between an LM's induced representation of a concept (e.g., the space of word embeddings for colors) and its physical (or human perception-like) representation (e.g., RGB space) has direct implications for how much explicit grounding is necessary for an LM to learn about the ``real world'' referred to by the textual data on which it was trained. And inasmuch as grounding may be relevant for meaning and understanding,\footnote{See \citet{pavlick2023symbols} and \citet{sogaard2023grounding}  for further discussion on the relationship between grounding and meaning.} these findings also have indirect implications for whether LMs can acquire (some operationalization of) meaning through text-only training \cite{bender_climbing_2020}.

This work investigates the extent to which LMs encode perceptual representations of \emph{sounds}. Past works have found that  LM representations of some objects are partially isomorphic to representations of those same objects from vision models \cite{ilharco_probing_2021,li_implications_2023}, suggesting that LMs are able to learn nontrivial structures about the visual world through just text-only training. We extend this setup to sounds through the lens of probing \cite{belinkov-2022-probing}, where we learn simple linear transformations that align the language representation for an object $c$  to its sound representation (from a pretrained audio model). If this retrieval-based probe generalizes to objects that were not seen during training, this suggests that there are structural similarities between the language and sound representations, i.e., LMs have learned meaningfully grounded representations of $c$ despite being just trained on raw text.

\begin{figure*}[t!]
  \vspace{-4mm}
    \centering
    \includegraphics[scale=0.65]{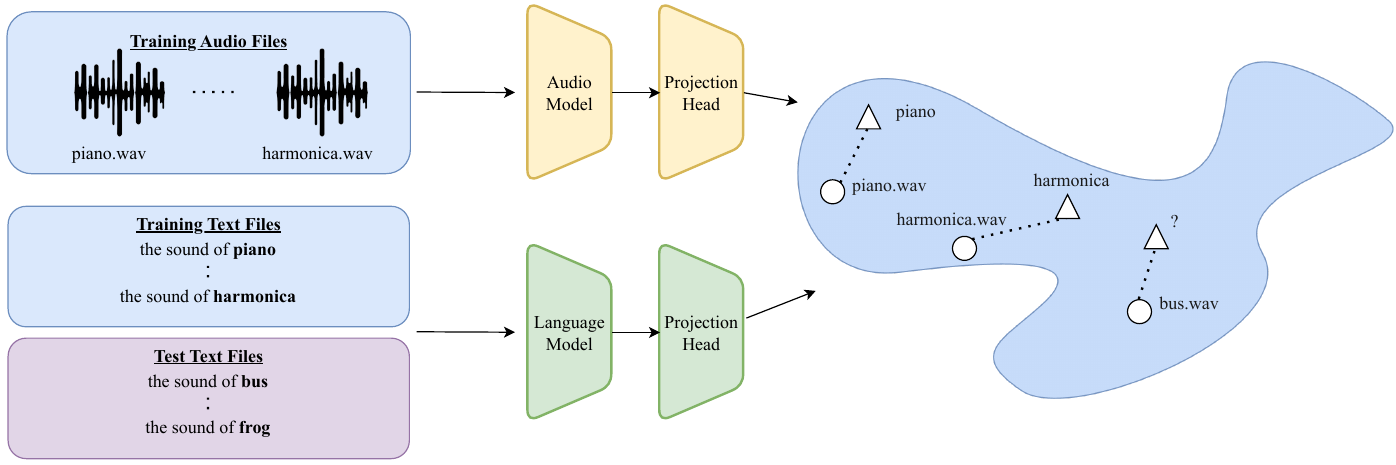}
       
\vspace{-1mm}
       \caption{An overview of our experimental setup. We randomly split a set of classes into mutually exclusive train/test sets. On the training set (blue), we use a contrastive loss to  learn linear transformations (i.e., projection heads) of the sound and language representations such that a language representation of a class is close in cosine distance to the sound representation of the same class. After training, we apply the learned probe on audio snippets of classes from the test set, and retrieve the most similar text representation (from classes in both the train and test sets). We then test whether the retrieved class corresponds to the actual class. }
       \vspace{-2mm}
    \label{fig:main_diagram}
\end{figure*}

We conduct the sound probing study across 6 language  models and 3 audio models. The language representations include those from word vector-only models (GloVe \citep{pennington-etal-2014-glove}, word2vec \citep{NIPS2013_9aa42b31}), encoders (BERT \citep{devlin-etal-2019-bert}, T5 \citep{2020t5}), and decoders (GPT-2 \citep{radford2019language}, LLaMA \citep{touvron2023llama}). On the audio side, we  experiment with two types of models: self-supervised models that have been pretrained without access to any external labels \citep[AudioMAE;][]{audiomae}, and supervised models that have been pretrained on sound event classification (finetuned AudoMAE, PANN 
 \citep{pann}, PaSST \citep{passt}). While all audio models are trained without explicit access to symbolic language data, the representations from supervised models implicitly encode more human perception-like priors given that the classification task itself incorporates information about  what snippet of sound constitutes a salient-enough signal to humans to warrant its being classified as a distinct event. That is, purely self-supervised models are more likely to encode more physical (i.e., acoustic) representations whereas supervised models are more likely to encode more human perception-like (i.e., auditory) representations. 
 
On both acoustic- and auditory-like sound representations, we find that all language models generalize to unseen classes at an above-chance level. We also find that the generalization performance is typically better for sound representations that have been supervised on sound event classification.

\section{Probing for Auditory Knowledge}
\label{sec:method}
\subsection{Preliminary Study: Procrustes Analysis}
\label{sec:proc}
We perform a preliminary qualitative study to test the feasibility of aligning language and sound representations. Let $\mathcal{C} = \{\texttt{car}, \texttt{bus}, \texttt{harmonica}, \texttt{harp} \dots \}$ be a set of objects. Further let $f_{\operatorname{LM}} : \Sigma^\ast \to \mathbb{R}^{d_1}$ be a text encoding function from a pretrained LM that produces a $d_1$-dimensional vector representation of a sentence (e.g., via averaging the contexualized word embeddings of $c$ occurring in some sentence $x \in \Sigma^{\ast}$), and similarly let $f_{\operatorname{AM}} : \mathbb{R}^{T \times m} \to \mathbb{R}^{d_2}$ be an audio encoding function from a pretrained audio model that takes in an $m$-dimensional audio signal\footnote{Where each dimension corresponds to a particular frequency in the case of spectrograms.} of (up to) length $T$ and produces a $d_2$-dimensional vector representation of that audio signal. 
For each $c \in \mathcal{C}$, let $\txt(c) \in \Sigma^\ast$ be the text template for $c$ that describes $c$'s sound,\footnote{Our template is ``\texttt{the sound of $c$}.''} and further let $\sound(c)$ be a sound associated with $c$ (e.g., the sound of a harmonica if $c = \texttt{harmonica}$). We are interested in comparing  the space of induced text representations $\mathcal{C}_{\operatorname{language}} = \{f_{\operatorname{LM}}(\txt(c)) : c \in \mathcal{C} \}$  with the sound representations $\mathcal{C}_{\operatorname{sound}} = \{f_{\operatorname{AM}}(\sound(c)) : c \in \mathcal{C} \}$; if the geometry of these representations is ``similar'' in some way (e.g., they are  isomorphic), then we can infer that the pretrained LM has nontrivial knowledge of sounds that are associated with objects in $\mathcal{C}$.\footnote{This is the case only if the pretrained audio model is trained without access to any symbolic language. This will indeed be the case in our experiments.}

As an initial study, we analyze these two spaces with Procrustes analysis. Let $\mathbf{C}_{\operatorname{language}} \in \mathbb{R}^{|\mathcal{C}| \times d_1}$ be the matrix obtained by stacking the text representations for each $c \in \mathcal{C}$, and similarly for $\mathbf{C}_{\operatorname{sound}} \in \mathbb{R}^{|\mathcal{C}| \times d_2}$.\footnote{In our dataset, there are multiple audio snippets associated with a single $c$. We average all audio embeddings associated with $c$ to obtain the object-level sound representation.} Since  we generally have $|\mathcal{C}| < d_2 < d_1$, we first perform PCA  on both matrices to obtain 
$\mathbf{C}_{\operatorname{language}}' \in \mathbb{R}^{|\mathcal{C}| \times |\mathcal{C}|}$ and 
$\mathbf{C}_{\operatorname{sound}}' \in \mathbb{R}^{|\mathcal{C}| \times |\mathcal{C}|}$. Procrustes analysis aligns these two matrices via minimizing
\begin{align*}
    \min_{\mathbf{Q} \in \mathbb{O}^{|\mathcal{C}| \times |\mathcal{C}|}} \,\, \Vert \mathbf{C}_{\operatorname{language}}'\mathbf{Q} - \mathbf{C}_{\operatorname{sound}}' \Vert^{2},
\end{align*}
where $\mathbb{O}^{|\mathcal{C}| \times |\mathcal{C}|}$ is the set of orthogonal matrices, i.e., $\mathbf{Q}^\top \mathbf{Q} = \mathbf{I}$. We perform PCA again to two dimensions and visualize the resulting representation space. Figure~\ref{fig:prelim} shows this analysis for BERT \cite{devlin-etal-2019-bert} and PaSST \cite{passt} on the FSD50K \cite{fonseca2021fsd50k} dataset. While far from perfectly aligned, there are reasonable symmetries. This motivates a more controlled, quantitative study described below.

\subsection{A Contrastive Probe}

While the above study indicates that there may be structural similarities between the two spaces, Procrustes analysis makes strong assumptions about the underlying geometry of the two spaces, which may be overly restrictive. We now describe a contrastive probe that is more flexible than the Procrustes ``probe'', which will be trained on a set of held-in objects and and tested on how it generalizes to a set of held-out objects.

Our probe uses the following (learned) similarity function between the language and sound representations based on cosine similarity,
\begin{align*}
    \simm&(\txt(c), \sound(c)) = \\
    &\frac{\langle \mathbf{W}_1 f_{\operatorname{LM}}(\txt(c)), \mathbf{W}_2 f_{\operatorname{AM}}(\sound{c})\rangle}{\left\Vert\mathbf{W}_1 f_{\operatorname{LM}}(\txt(c))\right\Vert \left\Vert\mathbf{W}_2 f_{\operatorname{AM}}(\sound{c})\right\Vert},
\end{align*}
with learnable linear transformations $\mathbf{W}_{1} \in \mathbb{R}^{d \times d_1}, \mathbf{W}_{2} \in \mathbb{R}^{d \times d_2}$ that project the text and audio embeddings into a common space. The above transformations can be learned in various ways; in this paper we use the standard contrastive loss objective,
\begin{align*}
    {L}&(\mathcal{C}) = \sum_{c \in \mathcal{C}} \Big( - \simm(\txt(c), \sound(c))/ \tau + \\
    &\log \sum_{c' \in N(c)} \exp \left(\simm(\txt(c), \sound(c'))/\tau \right)\Big),
\end{align*}
where $N(c)~\subseteq~\mathcal{C}~\setminus~\{c\}$ is a set of randomly sampled negative samples and  $\tau > 0$ is a temperature term.

Suppose we partition $\mathcal{C}$ into mutually exclusive train and test sets, $\mathcal{C} = \mathcal{C}^{\textrm{train}} \cup  \mathcal{C}^{\textrm{test}}$. If we learn $\mathbf{W}_1, \mathbf{W}_2$ via minimizing $L(\mathcal{C}^{\textrm{train}})$ and these transformations generalize to $\mathcal{C}^{\textrm{test}}$, i.e., for $c \in \mathcal{C}^{\textrm{test}}$
\begin{align*}
  c =   \arg\max_{\hspace{-4mm} c' \in \mathcal{C}} \,\, \simm(\txt(c'), \sound(c)),
\end{align*}
then this would suggest that the language representation space $\mathcal{C}_{\operatorname{language}}$ is structurally similar to the sound representation space $\mathcal{C}_{\operatorname{sound}}$. (Note that the $\arg\max$ is over the entire set $\mathcal{C}$). We thus use accuracy@$K$ over $\mathcal{C}^{\textrm{test}}$, where a prediction as counted as correct of the correct label is in the set of top $K$ most similar objects, to evaluate the alignment between a language model $f_{\operatorname{LM}}$ and an audio model $f_{\operatorname{AM}}$. See Figure~\ref{fig:main_diagram} for an overview.

\section{Experimental Setup}

\subsection{Models }
\paragraph{Language representations.}
We test representations from a variety of text models, including word2vec \cite{NIPS2013_9aa42b31}, GloVe \cite{pennington-etal-2014-glove}, BERT \cite{devlin-etal-2019-bert}, GPT-2 \cite{radford2019language}, T-5 \cite{2020t5}, and LLaMA \cite{touvron2023llama}. We also include several model versions within a family. 

To extract the language representations from the above models for a given object $c$, we obtain a sentence using the template ``\texttt{the sound of $c$}'' and average the contextualized representations for the tokens corresponding to $c$ within the resulting sentence.\footnote{For word2vec and GloVe,  we just average the representation of  ``$c$''.} Prior work on aligning language representations to visual representations have made use of natural sentences containing $c$ \cite{ilharco_probing_2021,li_implications_2023}. In order to eliminate confounding factors that may arise from  the other tokens in a sentence, we went with a simple templated approach for extracting the language representations.

\paragraph{Sound representations.} We test audio embeddings from three models:
AudioMAE \citep{audiomae}, PaSST \cite{passt} and PANN \citep{pann}. 

{AudioMAE} is a transformer-based model trained as a masked autoencoder on audio spectrograms. AudioMAE is pretrained via self-supervision on the AudioSet dataset \cite{gemmeke2017audio}, which contains approximately 2 million segments of audio snippets from YouTube along with annotated labels that describe the sound event of the audio snippet (e.g., \texttt{dog}, \texttt{cat}, \texttt{aircraft}, ...).\footnote{\label{footnote} Since the AudioSet dataset contains examples of human speech, AudioMAE's sound representation is arguably not completely independent of language. The set of possible labels for human speech snippets in the dataset are \{\texttt{male speech}, \texttt{female speech}, \texttt{child speech}, \texttt{conversation}, \texttt{narration}, \texttt{babbling}, \texttt{speech synthesizer}\}.} Self-supervised AudioMAE is trained only on the spectrogram inputs of AudioSet. We also experiment with a supervised, finetuned version of AudioMAE (AudioMAE-FT) that is finetuned as an audio classification model on the same AudioSet dataset. 

{PaSST} is also a spectrogram-based transformer model that has been trained as an audio classification model to predict sound events. PaSST is initialized from a vision transformer pretrained on ImageNet, which was shown to improve performance despite the difference in modalities \cite{gong2021ast}.\footnote{Thus PaSST representations potentially encode even more human perception-like priors.} PaSST performs two stages of supervised training: large-scale supervised learning on the broad AudioSet dataset, followed by smaller-scale finetuning on the FSD50K dataset \cite{fonseca2021fsd50k}, which is another (more freely-licensed) sound event classification dataset that inherits AudioSet's ontology.

Finally, PANN is a CNN-based model that is trained with supervision on the AudioSet dataset.
Unlike the above models whose input is in the frequency domain (i.e., spectrograms), PANN operates directly over the time domain. We use the CNN14 version of PANN.

\begin{figure*}[t!]
\centering
\vspace{-2mm}
\begin{subfigure}{0.48\textwidth}

    \centering
    \includegraphics[scale=0.22]{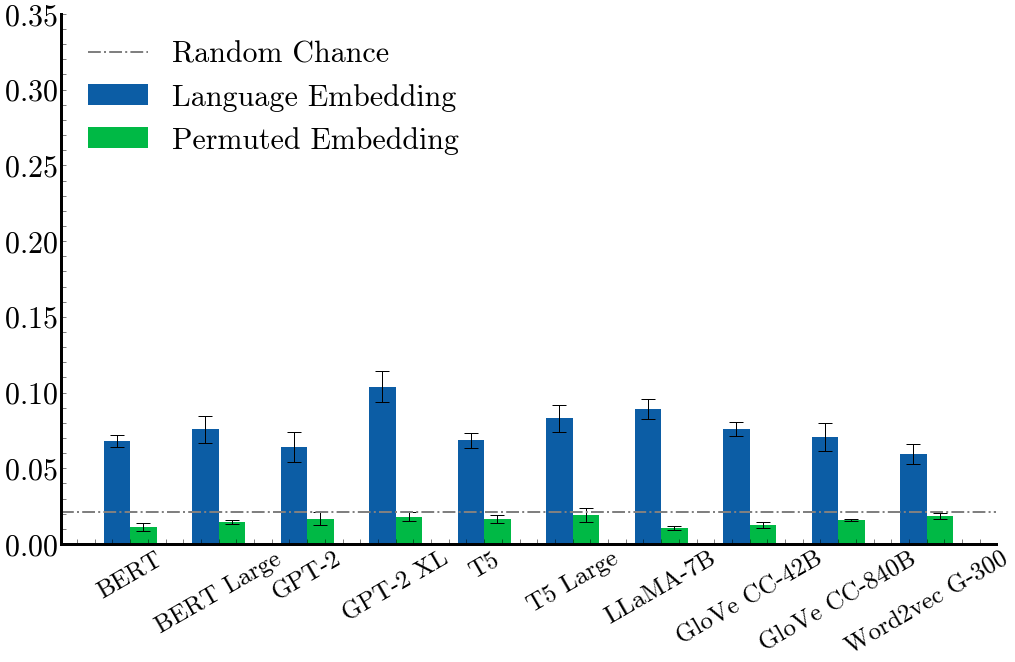}
    \label{fig:first}
        \vspace{-6mm}
    \caption{AudioMAE}
\end{subfigure}
\begin{subfigure}{0.48\textwidth}
    \centering
    \includegraphics[ scale=0.22]{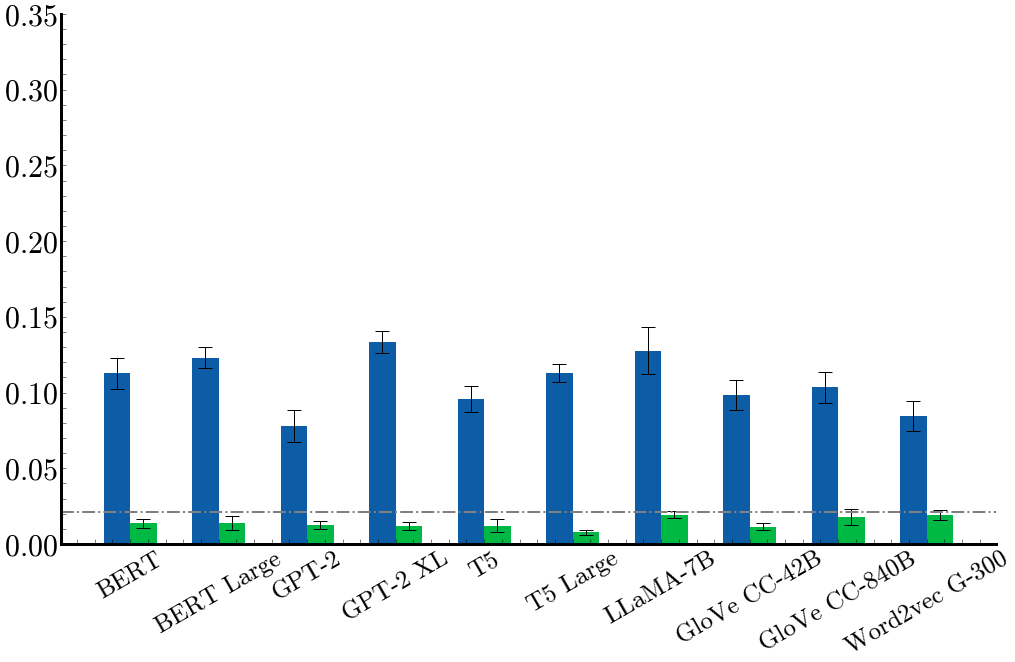}
    \label{fig:second}
        \vspace{-6mm}
    \caption{AudioMAE-FT}
\end{subfigure}
    \vspace{4mm}
\begin{subfigure}{0.48\textwidth}
    \centering
    \includegraphics[ scale=0.22]{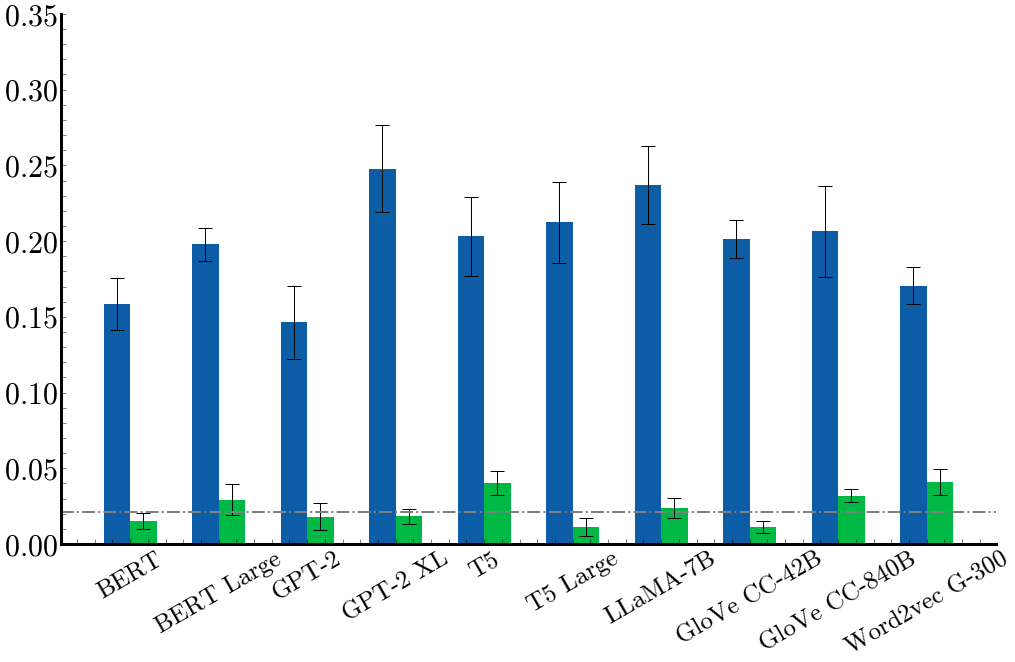}
    \label{fig:third}
        \vspace{-6mm}
    \caption{PaSST}
\end{subfigure}
\begin{subfigure}{0.48\textwidth}
    \centering
    \includegraphics[ scale=0.22]{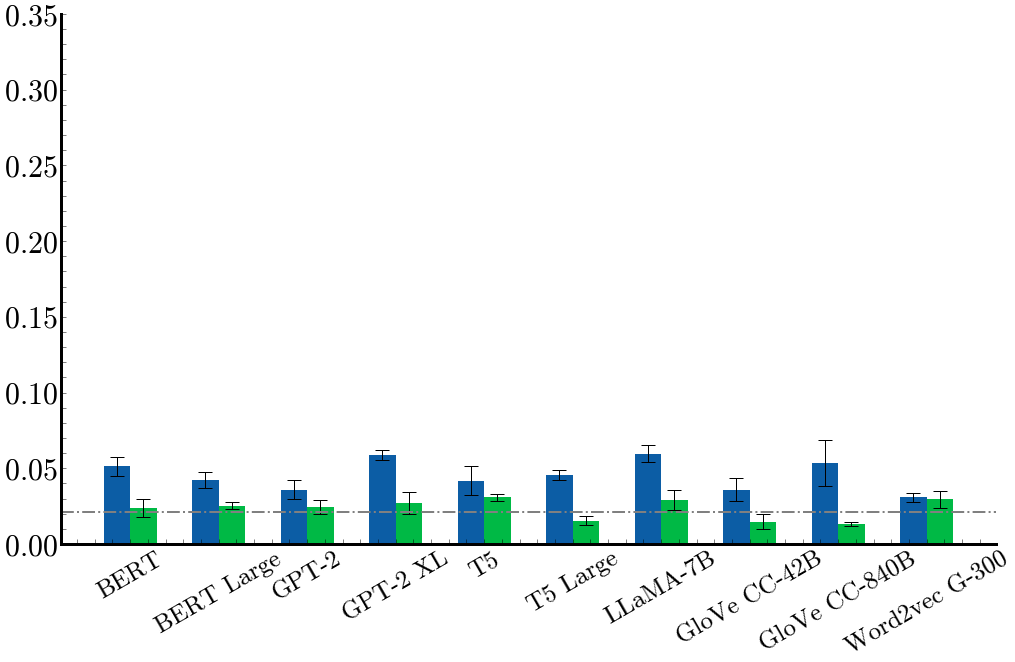}
    \label{fig:fourth}
    \vspace{-6mm}
    \caption{PANN}
\end{subfigure}
\vspace{-6mm}
\caption{Accuracy@3 for the different language and sound representations. Green bars show the accuracy of the permuted embedding control task, where the text representations are randomly permuted. Error bars show standard error of the mean across 5 runs. Dotted line shows random chance performance, which is 2.08\%.}
    \label{fig:graph}
\end{figure*}

\subsection{Dataset}
\label{sec:data}
Our main probing experiments are conducted on the FSD50K dataset \cite{fonseca2021fsd50k}, which includes around 50K audio clips with their annotated sound event classes with lengths ranging from 0.3-30 seconds. As some of the classes only have a few examples, we select the top 100 most frequent classes. In this case, each class has at least 117 audio samples.

Out of 100 classes, we randomly select 70 classes as the training object set $\mathcal{C}^{\text{train}}$, and learn the linear transformations $\mathbf{W}_1$ and $\mathbf{W}_2$ via the contrastive loss as described in \S\ref{sec:method}.\footnote{While the contrastive loss is presented assuming a single sample for $\sound(c)$, in practice we train over  multiple audio examples for a single object, instead of averaging the audio representations to obtain a single sound representation as was done in the Procrustes analysis (\S\ref{sec:proc}).} We then apply the probe on the 30 held-out classes $\mathcal{C}^{\text{test}}$ to obtain an accuracy@$K$ metric. For each audio snippet associated with $c$, prediction via retrieval is done over a superset of $\mathcal{C}$, in particular the set of 144 most frequent classes that were part of FSD50K (i.e.,  the retrieval set is over classes that were even outside the training set).  This increases the difficulty of the task, and a similar approach was adopted in the context of aligning text and vision representations \cite{li_implications_2023}.  We repeat this training and testing over  5 different random partitions of $\mathcal{C}$ (each with a 70/30 split), and report the average accuracy@3 (over 144 classes) for audio snippets in the  test set.

\subsection{Hyperparameters}
We performed a light grid search over the hyperparameters, in particular the learning rate $\alpha \in \{10^{-3}, 10^{-4}\}$, temperature coefficient $\tau \in \{ 0.07, 0.2\}$, number of negative samples $N(c) \in \{64, 128\}$. We use a batch size of 32 and train for 20 epochs.
Importantly, we found the optimal hyperparameters (and the optimal epoch for early stopping) based on a \textit{held-in} validation set that was randomly sampled from the training set. That is, none of the hyperparameters were tuned based on held-out performance on $C^\textrm{test}$.

\subsection{Control Task}
Because we evaluate accuracy only on the set of held-out objects, the usual caveats associated with probing (i.e., whether the above-chance performance is due to an LM's representations' meaningfully encoding the phenomena in question, or due to the probe's learning the task) are less of an issue. However, there may be other factors that may be contributing to above-chance performance, for example, the overall geometry of the respective representation spaces. We thus follow \citet{hewitt-liang-2019-designing} and also compare the performance of our probes against a control task where we randomly permute the text embeddings.

\begin{figure*}[t!]
\vspace{-6mm}
\centering
\begin{subfigure}{0.48\textwidth}
    \centering
    \includegraphics[scale=0.18]{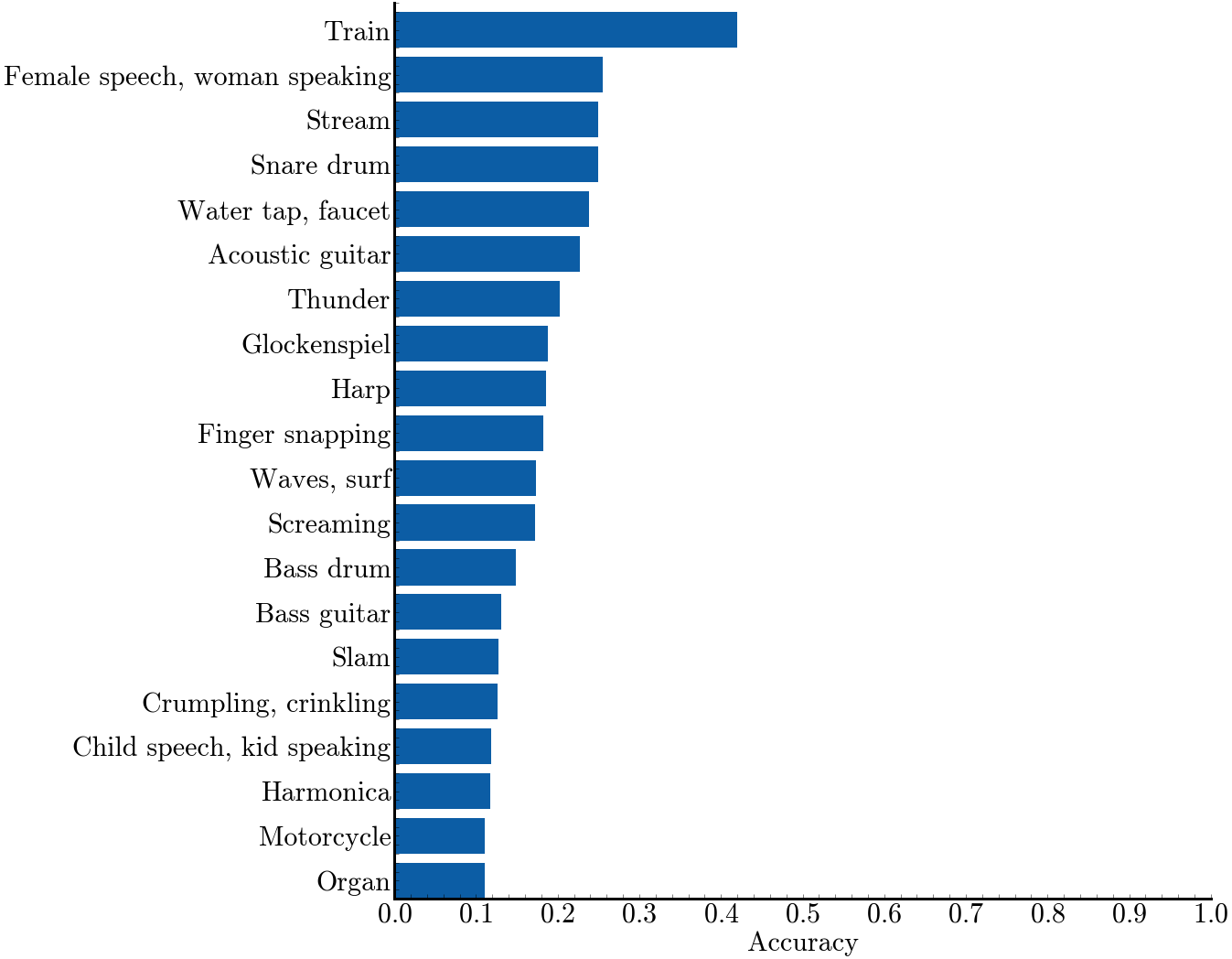}
    \vspace{-6mm}
    \caption{AudioMAE}
    \label{fig:first}
\end{subfigure}
\begin{subfigure}{0.48\textwidth}
    \centering
    \includegraphics[ scale=0.18]{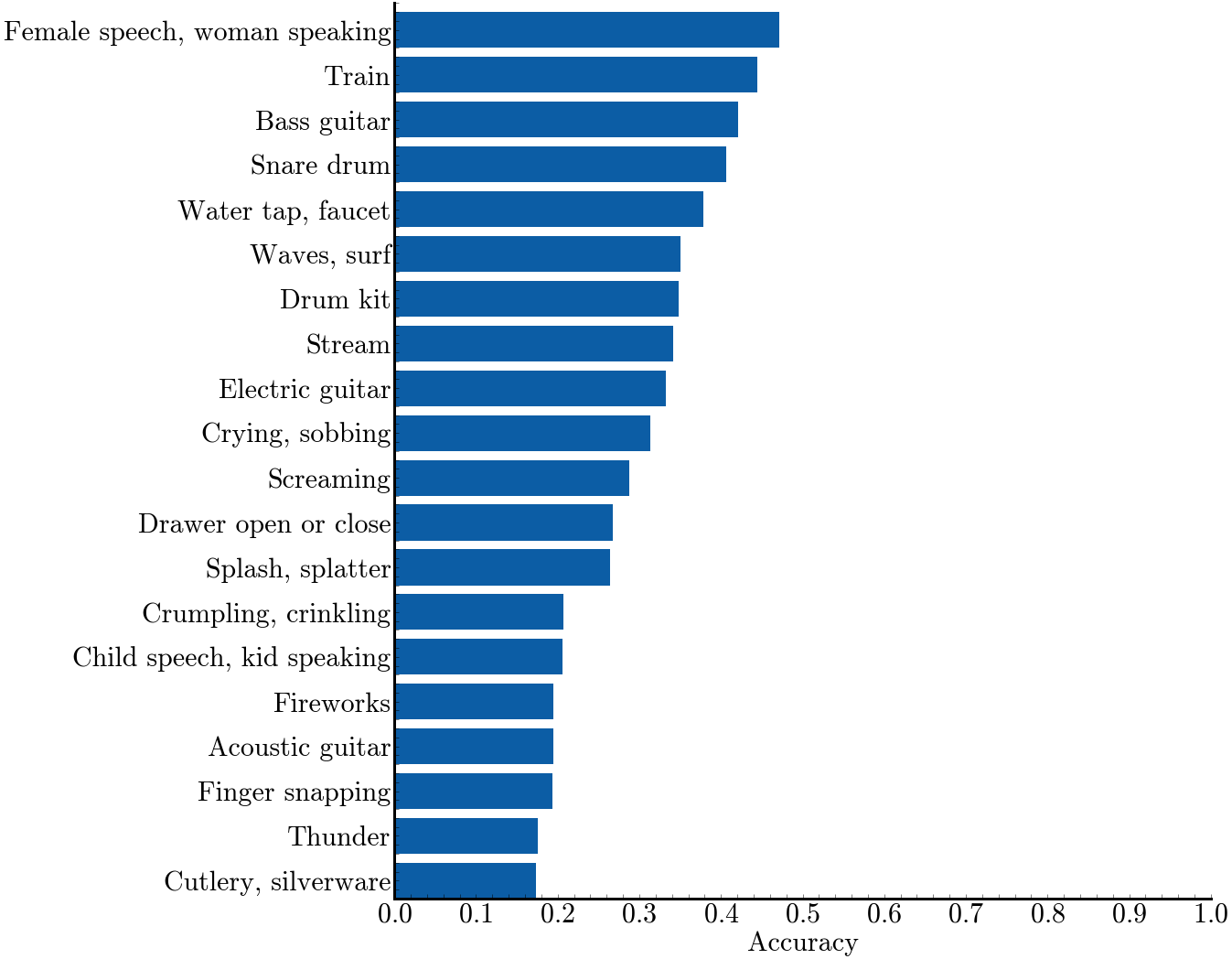}
        \vspace{-6mm}
    \caption{AudioMAE-FT}
    \label{fig:second}
\end{subfigure}

\vspace{4mm}
\begin{subfigure}{0.48\textwidth}
    \centering
    \includegraphics[ scale=0.18]{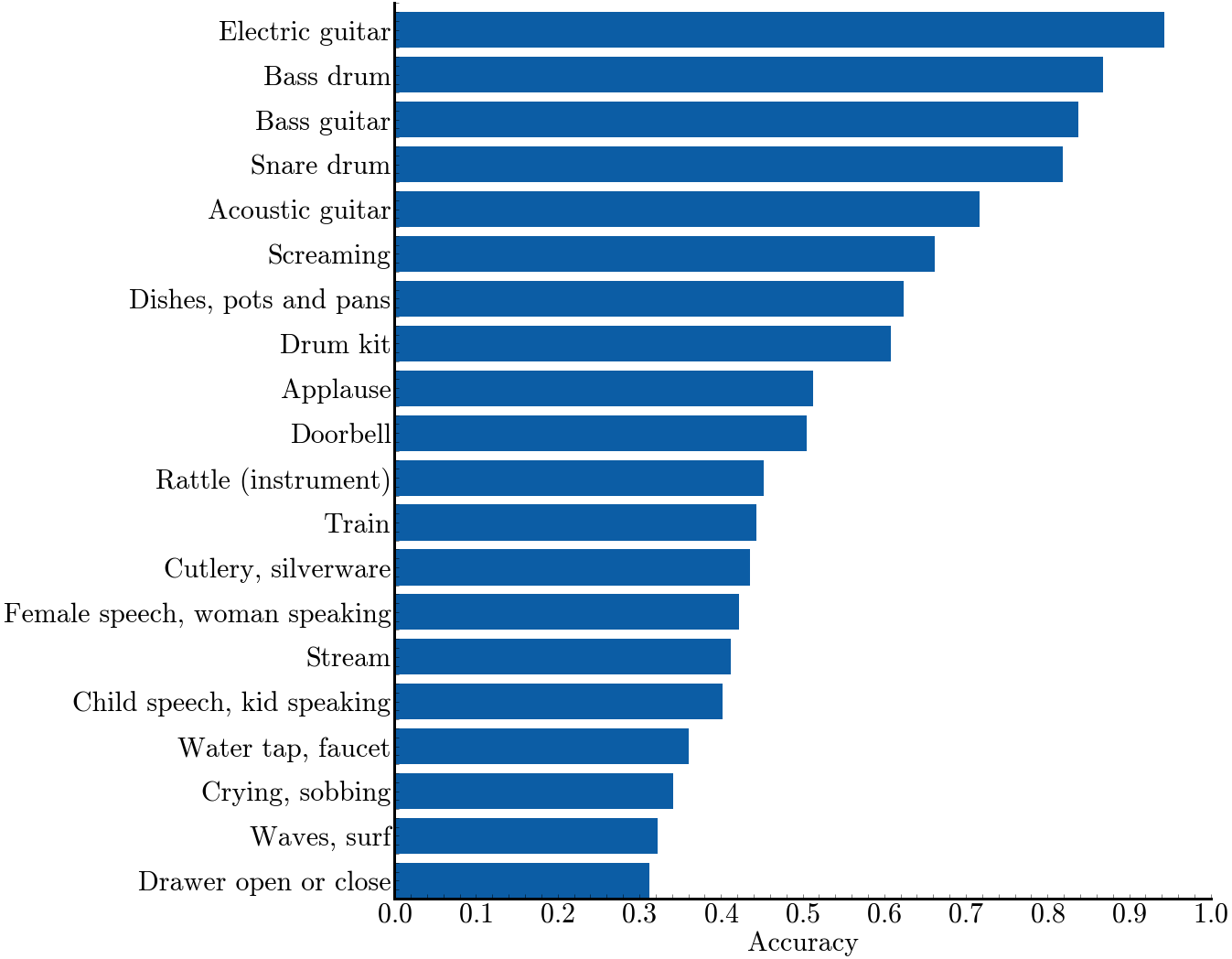}
    \caption{PaSST}
    \label{fig:third}
\end{subfigure}
\begin{subfigure}{0.48\textwidth}
    \centering
    \includegraphics[ scale=0.18]{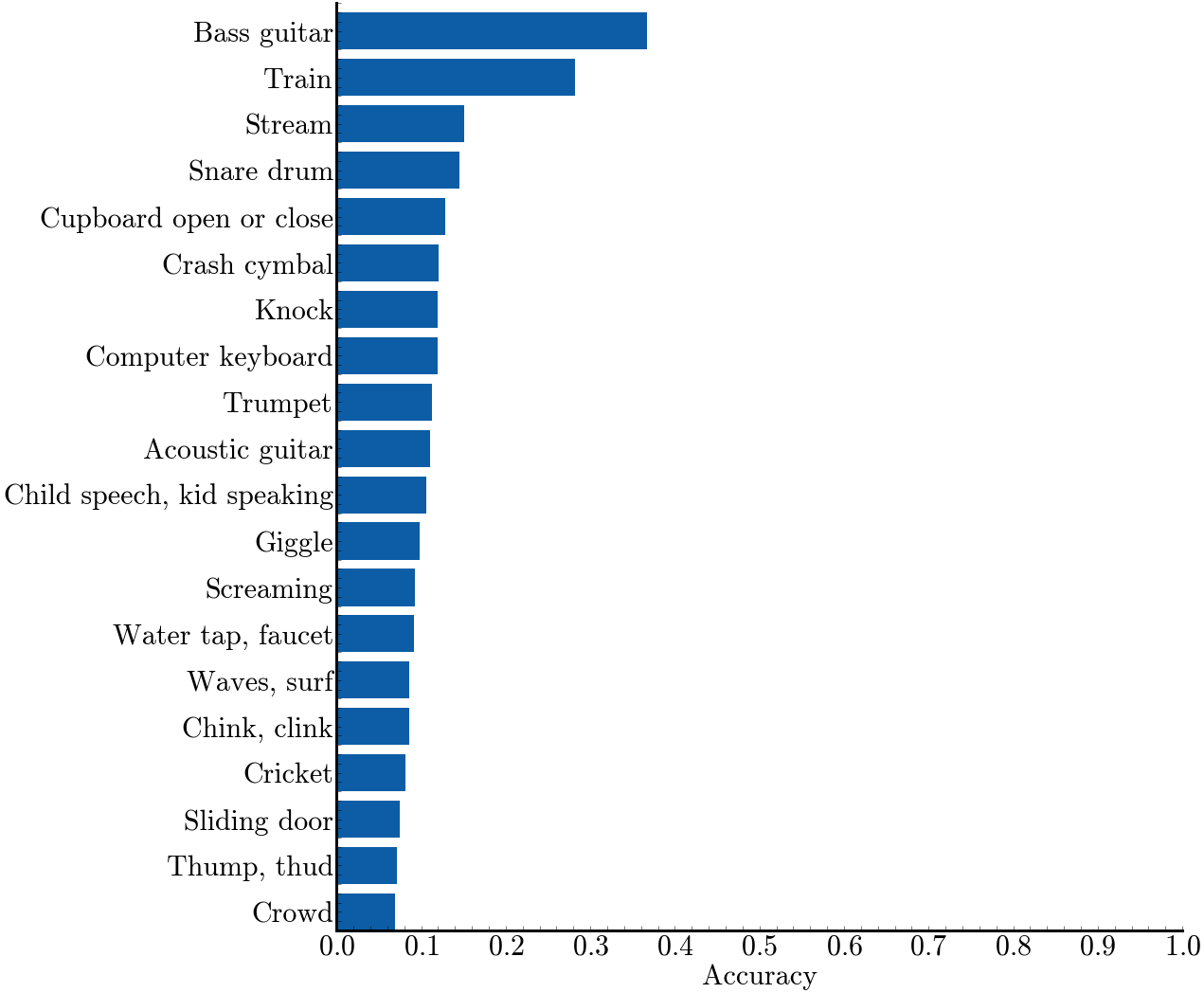}
    \caption{PANN}
    \label{fig:fourth}
\end{subfigure}
\vspace{-2mm}
\caption{Classes that had the best accuracies (as measured by accuracy@3) for the different sound representations. We measure the accuracies across all 5 train/test sets, and average across the different language models.}
    \label{fig:class_acc}
\end{figure*}

\begin{table*}
\scriptsize
\centering
\begin{tabular}{lrrllllll}
\toprule
& & & \multicolumn{3}{c}{\textbf{3 closest classes in language space}} & \multicolumn{3}{c}{\textbf{3 closest classes in sound space}} \\
  \textbf{OOD Classes} &    \textbf{Acc@1} &    \textbf{Acc@3} &          \textbf{Class 1} &          \textbf{Class 2} &          \textbf{Class 3} &         \textbf{Class 1} &         \textbf{Class 2} &      \textbf{Class 3} \\
\midrule
\textbf{AudioMAE} \\
\hspace{2mm} Female spee.. &   0.75 &   0.93 &   Male speech &          Yell &    Whispering &   Male speech &          Yell & Whispering \\
 \hspace{2mm}  Bass guitar &   0.19 &   0.78 & Acoustic gu.. & Electric gu.. &    Snare drum & Acoustic gu.. & Bowed strin.. &   Drum kit \\
\hspace{2mm}        Stream &   0.09 &   0.64 &         Waves &     Water tap &          Sink &     Water tap &         Waves &       Sink \\
\hspace{2mm}     Fireworks &   0.04 &   0.55 &       Thunder &       Gunshot & Keys jangli.. &       Gunshot &       Thunder &     Hammer \\
   \hspace{2mm}    Burping &   0.04 &   0.34 &       Chewing &     Livestock &          Fart &       Chewing &        Rattle &  Livestock \\
\hspace{2mm}     Harmonica &   0.02 &   0.23 &         Organ &       Marimba &      Ringtone & Wind instru.. &    Snare drum &      Organ \\

\midrule
\textbf{AudioMAE-FT} \\
\hspace{2mm} Female spee.. &   0.72 &   0.97 &   Male speech & Child speec.. &       Yell & Male speech &          Yell &     Screaming \\
  \hspace{2mm} Bass guitar &   0.22 &   0.61 & Electric gu.. &      Drum kit & Snare drum &    Drum kit & Electric gu.. & Acoustic gu.. \\
       \hspace{2mm} Dishes &   0.06 &   0.30 &       Cutlery &        Hi-hat &    Shatter &     Cutlery &        Hi-hat &          Hiss \\
    \hspace{2mm} Bass drum &   0.06 &   0.15 &    Snare drum &      Drum kit &     Hi-hat &    Drum kit &    Snare drum &     Screaming \\
         \hspace{2mm}  Wind &   0.05 &   0.23 &         Waves &        Whoosh &    Thunder &       Waves &          Bark &       Thunder \\
       \hspace{2mm}  Giggle &   0.04 &   0.72 &        Crying &         Cough &  Screaming &      Crying &         Cough &     Screaming \\
\midrule
\textbf{PaSST} \\
\hspace{2mm}  Female spee.. &   0.71 &   0.98 & Male speech &          Yell & Child speec.. & Male speech &          Yell & Child speec.. \\
       \hspace{2mm}  Dishes &   0.25 &   0.90 &     Cutlery & Keys jangli.. &          Sink &     Cutlery & Keys jangli.. &        Rattle \\
\hspace{2mm}  Fixed-wing .. &   0.09 &   0.64 &      Subway &           Bus &         Train &      Subway &       Gunshot &           Bus \\
     \hspace{2mm}  Scissors &   0.05 &   0.45 &     Cutlery & Keys jangli.. &        Hammer &     Cutlery & Keys jangli.. &        Rattle \\
       \hspace{2mm}  Stream &   0.03 &   0.34 &        Drip &        Splash &         Waves &      Splash &         Waves &     Water tap \\
    \hspace{2mm} Bass drum &   0.02 &   0.91 &    Drum kit &    Snare drum & Acoustic gu.. &    Drum kit &    Snare drum &        Rattle \\
\midrule
\textbf{PANN}\\
\hspace{2mm}  Bass guitar &   0.35 &   0.87 & Acoustic gu.. & Bowed strin.. &         Piano &     Piano & Acoustic gu.. & Bowed strin.. \\
     \hspace{2mm}  Stream &   0.08 &   0.45 &         Waves &         Train &          Sink &     Waves &        Sawing &          Buzz \\
     \hspace{2mm}  Zipper &   0.02 &   0.06 &       Writing &       Cutlery &        Camera &     Crack &       Shatter &          Drip \\
       \hspace{2mm}  Harp &   0.00 &   0.00 & Glockenspie.. &         Piano &       Marimba &     Organ &         Chirp & Glockenspie.. \\
       \hspace{2mm} Wind &   0.00 &   0.00 & Wind instru.. &          Buzz &         Waves & Tick-tock &        Sawing &          Hiss \\
       \hspace{2mm} Walk &   0.00 &   0.00 &         Knock &           Run & Child speec.. &       Run &     Crumpling &          Hiss \\
\bottomrule
\end{tabular}
\vspace{-1mm}

\caption{For the GPT-2-XL probe we we show the top 6 classes for which accuracy was the highest for each audio representation (for a given data split). For each class (which has multiple audio snippets associated with the class), we show the three closest classes as measured by cosine similarity both in language representation space and in sound representation space.}

\label{tab:closest}
\end{table*}

\section{Results}
Figure~\ref{fig:graph} shows the accuracy@3 metric for the different language/sound representation combinations. In the appendix,  we report the full numeric values, including accuracy@1 and standard error across the 5 runs.  We find that most language models perform well above chance. Moreover, within a model family, larger models almost always outperform their smaller siblings (e.g., BERT-Large vs. BERT, GPT-2-XL vs. GPT-2, T5-Large vs. T5). However, there is significant variation across families and larger models aren't always better across models (e.g., GPT-2-XL vs. LLaMA-7b). Despite their simplicity, word2vec and GloVe obtain nontrivial performance, sometimes outperforming much more sophisticated models. 

Across the different audio models, we find that alignment is overall better for sound representations from PaSST, which arguably is most aligned to human perception insofar as it is pretrained as an image classifier, and then finetuned as a sound event classifier. The underperformance of AudioMAE (which is pretrained only on spectrograms via self-supervision and thus likely to focus only on acoustic information) against AudioMAE-FT (which is finetuned as a supervised classifier on top of AudioMAE and thus likely to additionally encode auditory---i.e., human perception-like---information)  further highlights the importance of human-like representations that emerge from learning to predict human-derived labels. However, despite being trained as a supervised model, PANN performs the worst. This may be due to the fact that PANN's audio input is in the time domain, as well as the fact that PANN uses a CNN architecture instead of a transformer.

\begin{figure*}[t]
\centering
   \vspace{-6mm}
\begin{subfigure}{0.48\textwidth}
    \centering
    \includegraphics[scale=0.15]{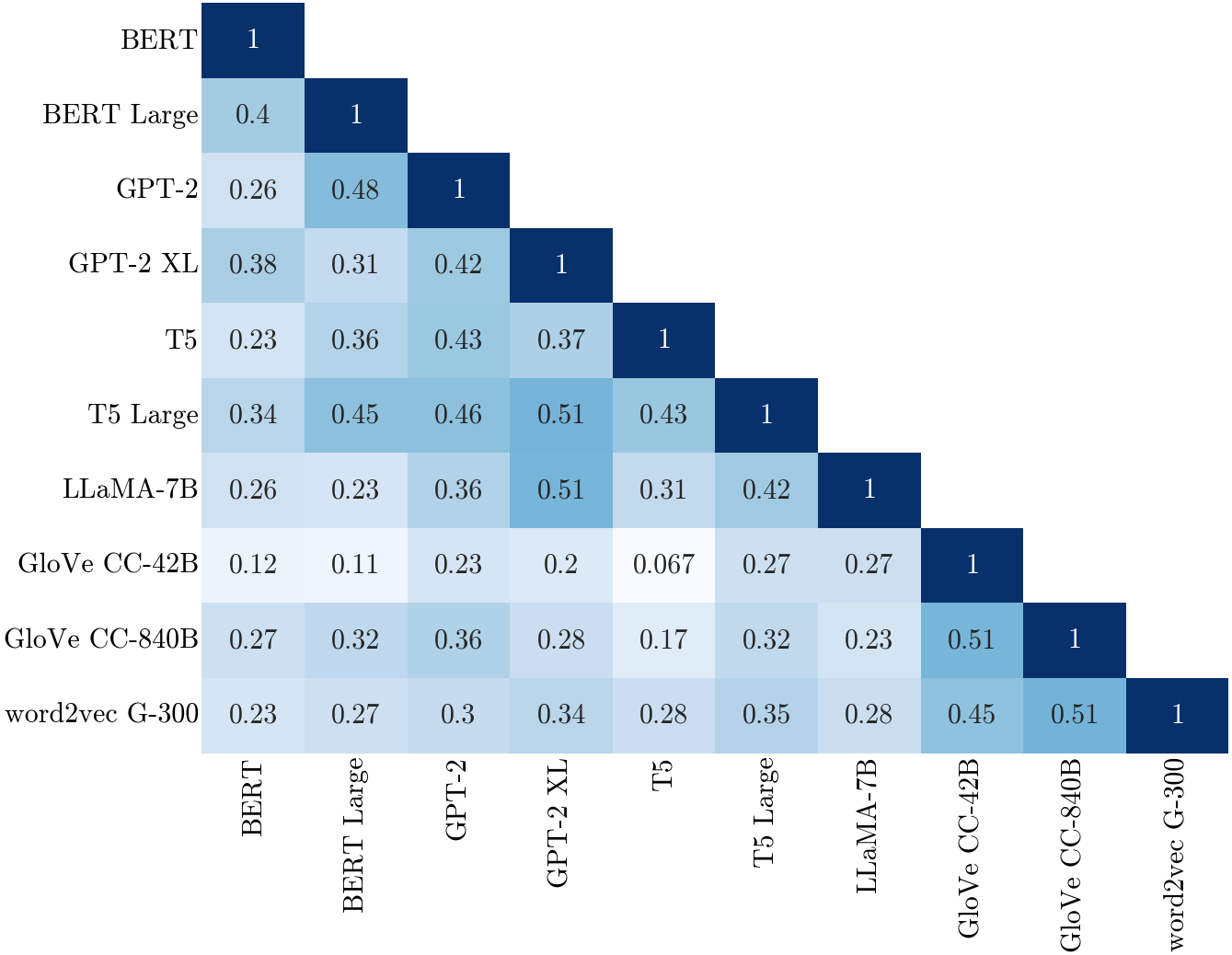}
    \vspace{-3mm}
    \caption{AudioMAE}
    \label{fig:first}
\end{subfigure}
\begin{subfigure}{0.48\textwidth}
    \centering
    \includegraphics[ scale=0.15]{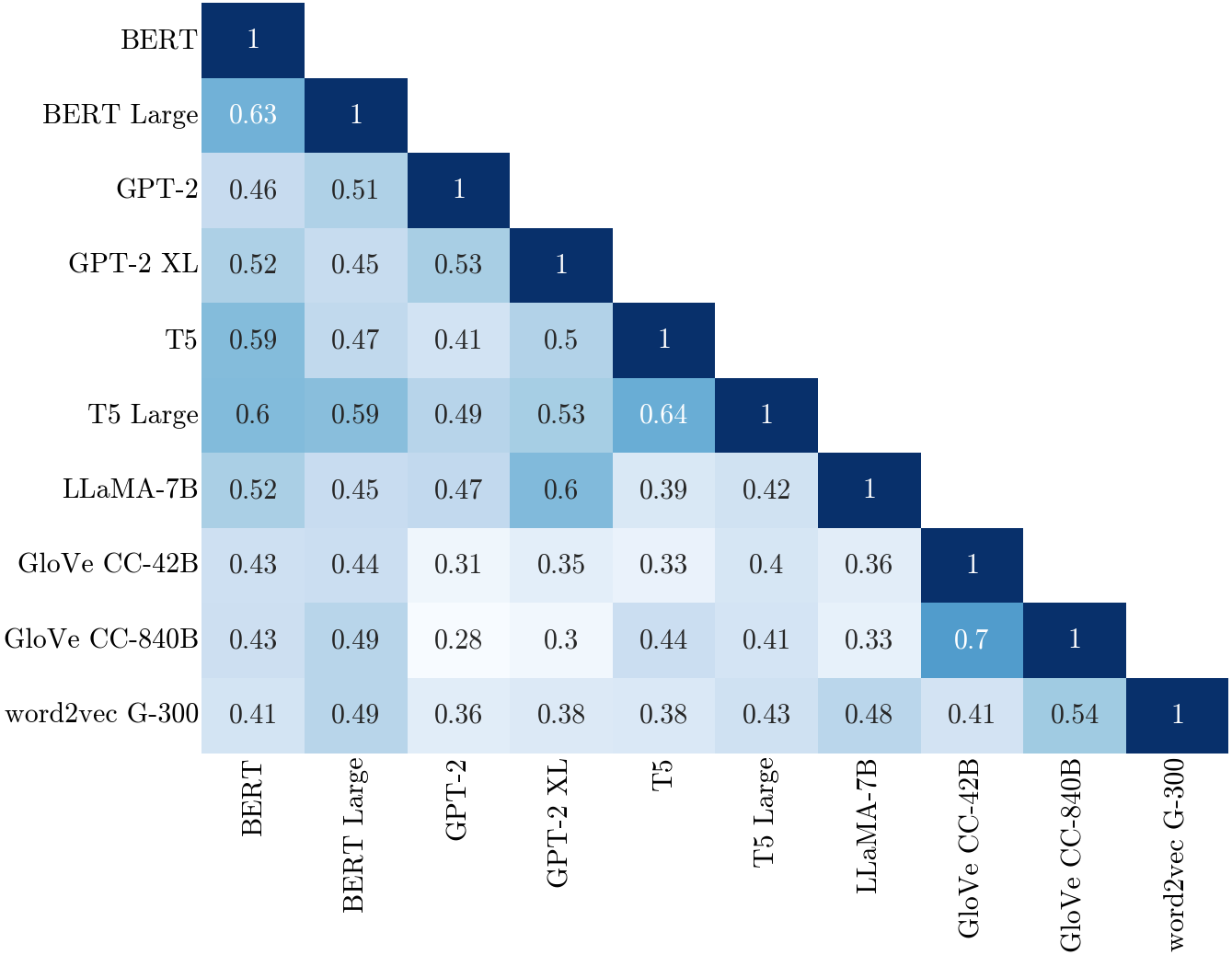}
        \vspace{-3mm}
    \caption{AudioMAE-FT}
    \label{fig:second}
\end{subfigure}

\vspace{8pt}
\begin{subfigure}{0.48\textwidth}
    \centering
    \includegraphics[ scale=0.15]{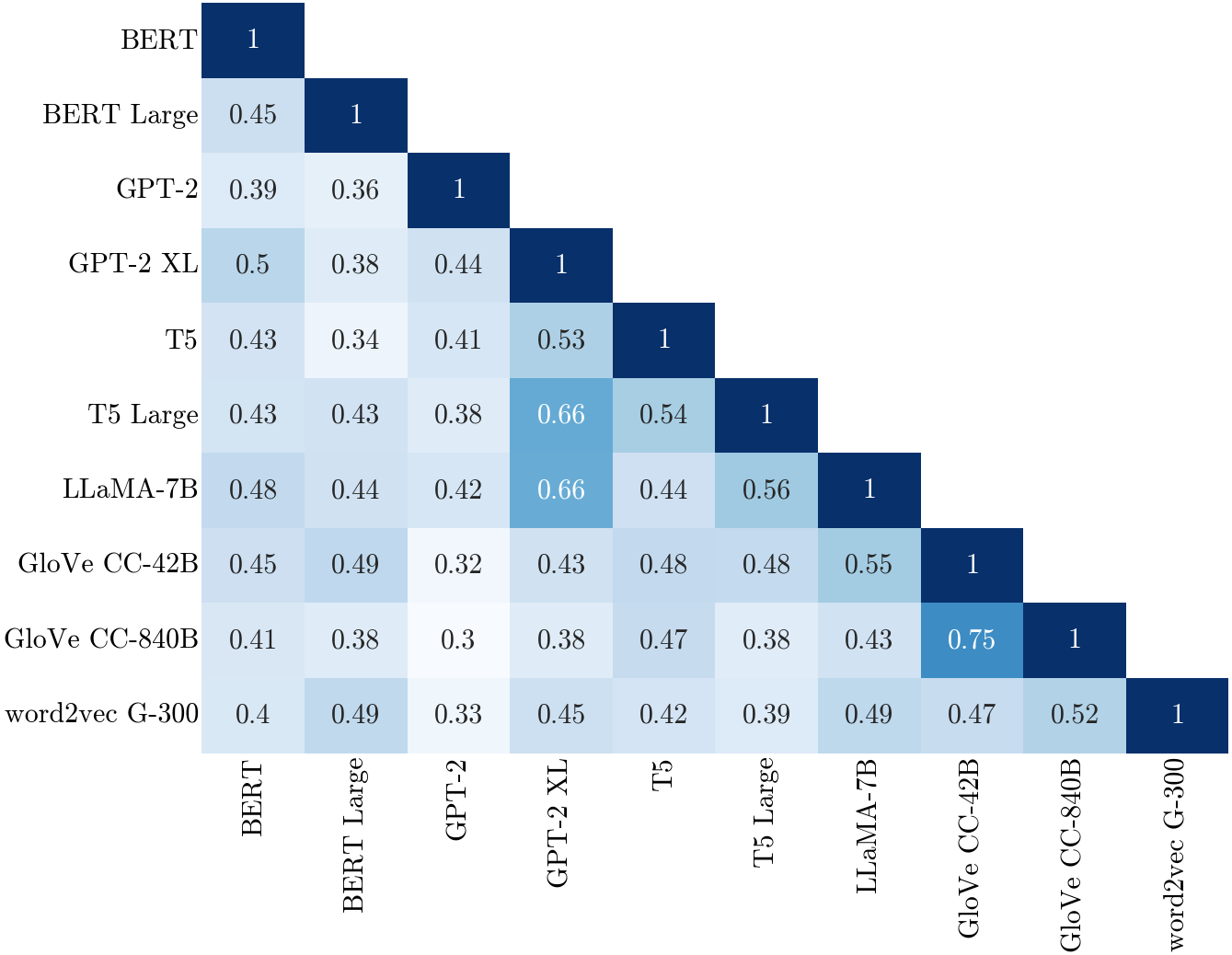}
        \vspace{-3mm}
    \caption{PaSST}
    \label{fig:third}
\end{subfigure}
\begin{subfigure}{0.48\textwidth}
    \centering
    \includegraphics[ scale=0.15]{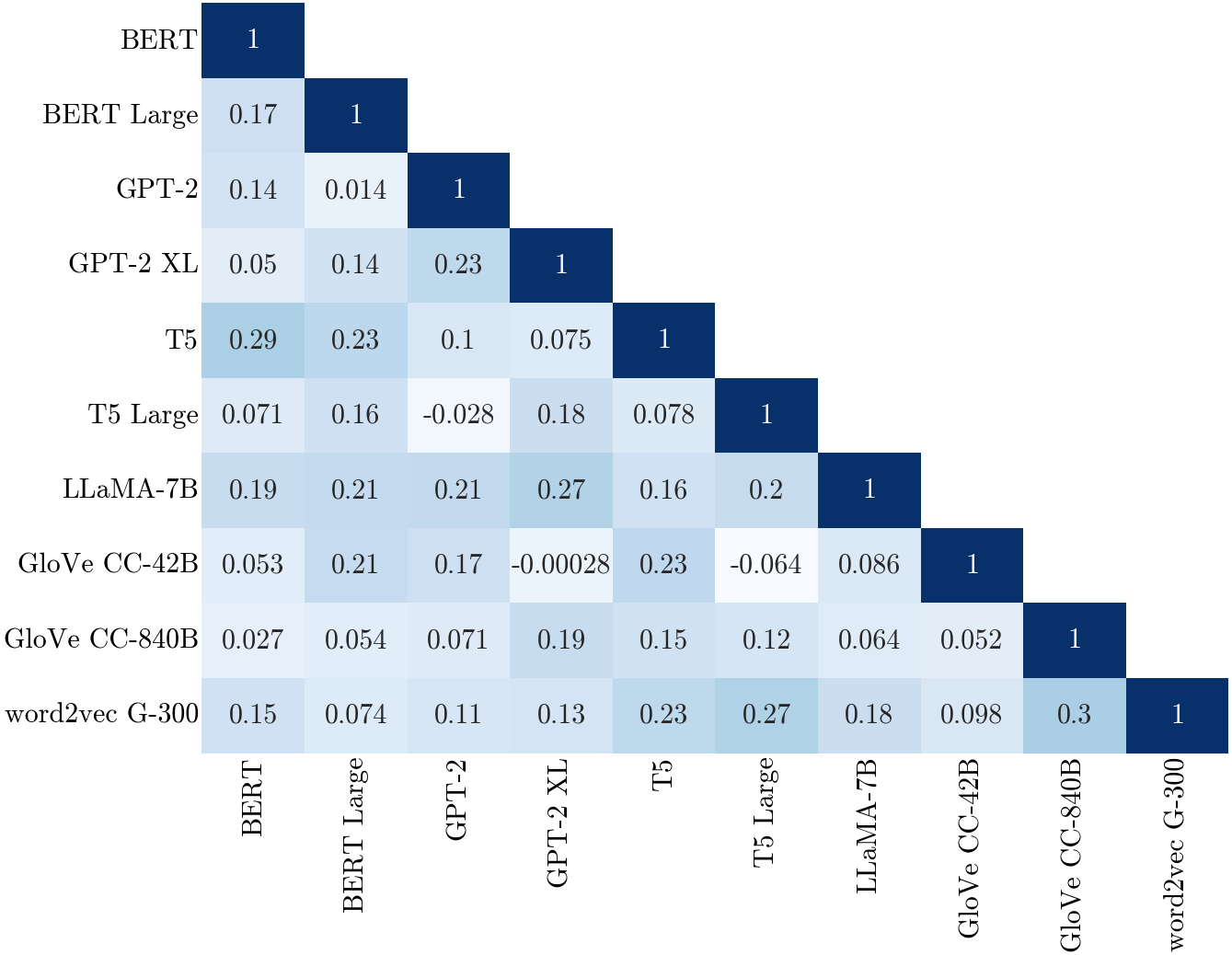}
        \vspace{-3mm}
    \caption{PANN}
    \label{fig:fourth}
\end{subfigure}
\vspace{-2mm}
\caption{Rank correlation of accuracies of classes within the test set between language representations, where the correlations are averaged across the five runs.}
    \label{fig:rank}
\vspace{-2mm}
\end{figure*}

\subsection{Analysis}
What are the classes for which the probes generalize particularly well? In Figure~\ref{fig:class_acc} we show the classes that had the best accuracies (averaged across the language representations) across the five runs. We qualitatively find that classes that correspond to human speech, as well as instruments, seem to generalize well. 

In Table~\ref{tab:closest} we perform a deeper analysis of GPT-2-XL, the best-performing language representation.  For each class in Table~\ref{tab:closest} (ranked by accuracy), we show the top 3 closest classes in the set of training classes as measured by similarity in language/sound space.\footnote{Note that training accuracies were extremely high (e.g., >97\%) for most classes. Therefore, it is not the case that (for example) ``\texttt{Electric guitar}'' accuracy is high because the text embedding for ``\texttt{Electric guitar}'' is the closest language embedding for \emph{all} audio snippets.} In many cases, the top 3 retrieved classes are similar in both spaces, indicating structural similarities.

We next analyze whether classes that obtain high accuracies are in general similar across the different language models. For two language representations, we calculate Spearman's rank correlation between the accuracies of classes in the test set. We average this rank correlation across the 5 runs, which produces a measure of how similar the two language representations are in terms of their ability to encode sound information. Figure~\ref{fig:rank} shows the results for all pairwise correlations. As expected, language representations are generally the most similar within a language model family, although this is not always the case. Correlation is generally quite high across the different language representations despite the differences in model architectures, size, and training data; this potentially implies that there is a common set of classes for which sound representation is meaningfully encoded.

\begin{table}[t]
\footnotesize
\centering
\begin{tabular}{lrrr}
\toprule
\textbf{Audio Model} &  \textbf{Linear} & \textbf{Non-Linear} &  \textbf{Procrustes} \\
\midrule
 AudioMAE &    0.11 &  0.10 &        0.09 \\
     AudioMAE-FT &    0.08 &  0.08 &        0.03 \\
     PaSST &    0.20 &  0.16 &        0.17 \\    
     PANN &    0.05 &  0.04 &        0.02 \\
\bottomrule
\end{tabular}

\caption{Generalization performance (accuracy@3) of different probes, where the performance is averaged across all language model representations.}
\label{fig:alignment_perform}
\end{table}

\subsection{Probing Method}
Our primary results make use of a contrastive loss with linear transformations applied to the language/sound representations. We additionally explore two other probes: the Procrustes probe discussed in \S\ref{sec:proc} where we learn the matrix $\mathbf{Q}$ only based on the objects in $\mathcal{C}^{\textrm{train}}$; and a non-linear probe where we apply a ReLU non-linearity after projecting when calculating the similarity function. 

The results are shown Table~\ref{fig:alignment_perform}. We generally find that Procrustes probes, which minimize the MSE and additionally constrain the transformations to be orthogonal, underperform the contrastive loss probes. The linear probe outperforms the non-linear probe. 

\vspace{-1mm}
\section{Discussion and Limitations}
\vspace{-1mm}

Our work, along with the line of work on aligning language model representations to grounded representations, provides evidence that modeling statistical correlations among surface-form text could lead to learning nontrivial structures about the real world.  In hindsight, this is perhaps not so surprising; both language and sound are different ``projections'' of the same physical world, and thus it is not inconceivable that models trained on the respective modalities represent (some) aspects of the original physical world in a similar way.

More generally, the fact that current language models (and foundation models more generally) are  trained only on ``raw form'' (such as word pieces, sound waves, pixels, etc.) is not an inherent limitation on their ability to learn physically grounded conceptual spaces. These models are typically trained (implicitly or explicitly) to compress their training data into their parameters; insofar as good compression can be achieved by learning the underlying generative process, it is possible that aspects of the physical world which were involved in the generation of language could be learned just through form-only training. Nonetheless,  form-only training is likely to be highly data-inefficient.

This work only explored whether sound representations that were {learned} by an auxiliary audio processing model were encoded through text. Here we found that sound representations that are more likely to encode auditory (i.e., human perception-like) information  were more aligned to the text representations than sound representations from purely self-supervised models which were just trained on spectrograms, and thus more likely to encode acoustic information. However, even the self-supervised audio models implicitly encode human perception-aligned priors given that the input data consisted of snippets of audio that corresponded to different sound events, which itself is derived from humans. It would be interesting to see whether it is possible to probe out even more low-level representations of objects (e.g., raw spectrograms, pixels) from language models. Similarly, as discussed in footnote~\ref{footnote} our audio representations are not completely independent of language as their training sets  included a significant amount of human speech. It would therefore be  interesting to see if audio models trained without any human speech learn representatons that can be aligned to language models.

\section{Related Work}
\paragraph{Probing language models.} Language models have been shown to encode much linguistic information in their contextualized representations \citep{tenney2019bert,liu-etal-2019-linguistic,jawahar2019does} and attention distributions \citep{clark2019does,vig2019analyzing}. Building on top of these more linguistically-oriented probes, there has been mounting recent evidence that language models trained on just text are able to meaningfully encode a surprising amount of grounded or extralinguistic information, such as color \cite{abdou-etal-2021-language}, direction \cite{patel2021mapping}, size \cite{zhang-etal-2020-language-embeddings,grand2022semantic}, geography \cite{konkol-etal-2017-geographical,lietard2021language,faisal-anastasopoulos-2023-geographic,chen2023more}, time \cite{gurnee2023language}, visual representations \citep{ilharco_probing_2021,merullo2022linearly,li_implications_2023}, character-level information of word-pieces \citep{kaushal-mahowald-2022-tokens}, and  representations of meaning \citep{li-etal-2021-implicit}.   LM-derived similiarity measures have also been shown to correlate with human-derived similarity measures across a number of modalities \cite{marjieh2023large,marjieh2022words}. The present work extends the probing-based line of work to sounds, and investigates the extent to which language models trained on text-only can encode auditory representations. 

Our work is also related to the line of work investigating whether a model that has been trained on raw outputs of a synthetic environment can acquire ``true'' representations of that environment. Examples of such environments include Othello \cite{li2022emergent}, chess \cite{toshniwal2021learning}, and toy grid worlds \cite{yun2023emergence,jin2023evidence}.

\vspace{-1mm}
\paragraph{Meaning  in language models.} Whether language models can acquire meaning and understanding from being trained on form alone is the subject of much debate \cite{bender_climbing_2020,merrill-etal-2021-provable,piantadosi2022meaning,pavlick2023symbols,sogaard2023grounding}.   In operationalizations  of meaning which do not rely on explicit reference to the external world, the fact that the geometry of language models' representation spaces is structurally related to the geometry of grounded representations could be construed as evidence for these models' acquiring meaning in some broad sense. 

\vspace{-1mm}
\section{Conclusion}
\vspace{-1mm}
We probe text-only language models for whether their representations of an object contain grounded representations of the sounds of the same object. We find that this is indeed the case, and a contrastive probe can often generalize zero-shot to object classes not seen during training.

\section*{Acknowledgments}
This study was partially supported by funds from the MIT-IBM Watson AI Lab.

\bibliography{anthology,custom}

\appendix

\section{Appendix}
We show the full numeric results for our main probing experiments in table~\ref{tab:full} and for classes accuracy in figure~\ref{fig:no_cut}.

\begin{table*}
\scriptsize
\centering
\begin{tabular}{lllllll}
\toprule
 & \multicolumn{2}{c}{\textbf{Language Embedding}} & \multicolumn{2}{c}{\textbf{Permuted Embedding}} & \multicolumn{2}{c}{\textbf{Random Init}} \\
\textbf{Models} & 
\multicolumn{1}{c}{\textbf{A@1}} & 
\multicolumn{1}{c}{\textbf{A@3}} & 
\multicolumn{1}{c}{\textbf{A@1}} & 
\multicolumn{1}{c}{\textbf{A@3}} & 
\multicolumn{1}{c}{\textbf{A@1}} & 
\multicolumn{1}{c}{\textbf{A@3}} \\ \midrule
\textbf{AudioMAE} \\
\hspace{2mm} BERT & 0.02 ± 0.0 & 0.07 ± 0.0 & 0.0 ± 0.0 & 0.01 ± 0.0 & 0.0 ± 0.0 & 0.01 ± 0.01 \\
\hspace{2mm} BERT Large & 0.02 ± 0.0 & 0.08 ± 0.01 & 0.0 ± 0.0 & 0.01 ± 0.0 & 0.0 ± 0.0 & 0.02 ± 0.01 \\
\hspace{2mm} GPT-2 & 0.01 ± 0.0 & 0.06 ± 0.01 & 0.0 ± 0.0 & 0.02 ± 0.0 & 0.0 ± 0.0 & 0.02 ± 0.01 \\
\hspace{2mm} GPT-2 XL & 0.02 ± 0.0 & 0.1 ± 0.01 & 0.0 ± 0.0 & 0.02 ± 0.0 & 0.01 ± 0.0 & 0.03 ± 0.01 \\
\hspace{2mm} T5 & 0.02 ± 0.0 & 0.07 ± 0.0 & 0.0 ± 0.0 & 0.02 ± 0.0 & 0.02 ± 0.0 & 0.03 ± 0.01 \\
\hspace{2mm} T5 Large & 0.02 ± 0.0 & 0.08 ± 0.01 & 0.01 ± 0.0 & 0.02 ± 0.0 & 0.0 ± 0.0 & 0.01 ± 0.0 \\
\hspace{2mm} LLaMA-7B & 0.02 ± 0.0 & 0.09 ± 0.01 & 0.0 ± 0.0 & 0.01 ± 0.0 & 0.01 ± 0.0 & 0.03 ± 0.0 \\
\hspace{2mm} GloVe CC-42B & 0.02 ± 0.0 & 0.08 ± 0.0 & 0.0 ± 0.0 & 0.01 ± 0.0 & 0.01 ± 0.0 & 0.03 ± 0.01 \\
\hspace{2mm} GloVe CC-840B & 0.02 ± 0.0 & 0.07 ± 0.01 & 0.0 ± 0.0 & 0.02 ± 0.0 & 0.0 ± 0.0 & 0.01 ± 0.0 \\
\hspace{2mm} word2vec GNews-300 & 0.01 ± 0.0 & 0.06 ± 0.01 & 0.0 ± 0.0 & 0.02 ± 0.0 & 0.0 ± 0.0 & 0.01 ± 0.0 \\
\midrule
\textbf{AudioMAE-FT} \\
\hspace{2mm} BERT & 0.02 ± 0.0 & 0.11 ± 0.01 & 0.0 ± 0.0 & 0.01 ± 0.0 & 0.0 ± 0.0 & 0.02 ± 0.0 \\
\hspace{2mm} BERT Large & 0.02 ± 0.01 & 0.12 ± 0.01 & 0.0 ± 0.0 & 0.01 ± 0.0 & 0.0 ± 0.0 & 0.02 ± 0.0 \\
\hspace{2mm} GPT-2 & 0.01 ± 0.0 & 0.08 ± 0.01 & 0.0 ± 0.0 & 0.01 ± 0.0 & 0.01 ± 0.0 & 0.02 ± 0.0 \\
\hspace{2mm} GPT-2 XL & 0.02 ± 0.0 & 0.13 ± 0.01 & 0.0 ± 0.0 & 0.01 ± 0.0 & 0.01 ± 0.01 & 0.03 ± 0.01 \\
\hspace{2mm} T5 & 0.01 ± 0.0 & 0.1 ± 0.01 & 0.0 ± 0.0 & 0.01 ± 0.0 & 0.0 ± 0.0 & 0.01 ± 0.0 \\
\hspace{2mm} T5 Large & 0.02 ± 0.0 & 0.11 ± 0.01 & 0.0 ± 0.0 & 0.01 ± 0.0 & 0.0 ± 0.0 & 0.01 ± 0.0 \\
\hspace{2mm} LLaMA-7B & 0.02 ± 0.0 & 0.13 ± 0.01 & 0.0 ± 0.0 & 0.02 ± 0.0 & 0.0 ± 0.0 & 0.01 ± 0.01 \\
\hspace{2mm} GloVe CC-42B & 0.02 ± 0.0 & 0.1 ± 0.01 & 0.0 ± 0.0 & 0.01 ± 0.0 & 0.0 ± 0.0 & 0.02 ± 0.0 \\
\hspace{2mm} GloVe CC-840B & 0.02 ± 0.0 & 0.1 ± 0.01 & 0.01 ± 0.0 & 0.02 ± 0.0 & 0.01 ± 0.0 & 0.03 ± 0.01 \\
\hspace{2mm} word2vec GNews-300 & 0.01 ± 0.0 & 0.08 ± 0.01 & 0.0 ± 0.0 & 0.02 ± 0.0 & 0.01 ± 0.0 & 0.01 ± 0.0 \\
\midrule
\textbf{PaSST} \\
\hspace{2mm} BERT & 0.02 ± 0.01 & 0.16 ± 0.02 & 0.0 ± 0.0 & 0.02 ± 0.0 & 0.01 ± 0.01 & 0.02 ± 0.01 \\
\hspace{2mm} BERT Large & 0.03 ± 0.01 & 0.2 ± 0.01 & 0.0 ± 0.0 & 0.03 ± 0.01 & 0.0 ± 0.0 & 0.02 ± 0.01 \\
\hspace{2mm} GPT-2 & 0.02 ± 0.0 & 0.15 ± 0.02 & 0.0 ± 0.0 & 0.02 ± 0.01 & 0.01 ± 0.01 & 0.03 ± 0.01 \\
\hspace{2mm} GPT-2 XL & 0.04 ± 0.01 & 0.25 ± 0.03 & 0.0 ± 0.0 & 0.02 ± 0.0 & 0.01 ± 0.01 & 0.01 ± 0.01 \\
\hspace{2mm} T5 & 0.02 ± 0.01 & 0.2 ± 0.02 & 0.01 ± 0.0 & 0.04 ± 0.01 & 0.01 ± 0.01 & 0.03 ± 0.01 \\
\hspace{2mm} T5 Large & 0.02 ± 0.01 & 0.21 ± 0.02 & 0.0 ± 0.0 & 0.01 ± 0.01 & 0.01 ± 0.01 & 0.01 ± 0.01 \\
\hspace{2mm} LLaMA-7B & 0.04 ± 0.01 & 0.24 ± 0.02 & 0.0 ± 0.0 & 0.02 ± 0.01 & 0.0 ± 0.0 & 0.01 ± 0.01 \\
\hspace{2mm} GloVe CC-42B & 0.02 ± 0.01 & 0.2 ± 0.01 & 0.0 ± 0.0 & 0.01 ± 0.0 & 0.01 ± 0.01 & 0.01 ± 0.01 \\
\hspace{2mm} GloVe CC-840B & 0.03 ± 0.01 & 0.21 ± 0.03 & 0.01 ± 0.0 & 0.03 ± 0.0 & 0.01 ± 0.01 & 0.01 ± 0.01 \\
\hspace{2mm} word2vec GNews-300 & 0.02 ± 0.0 & 0.17 ± 0.01 & 0.0 ± 0.0 & 0.04 ± 0.01 & 0.01 ± 0.01 & 0.01 ± 0.01 \\
\midrule
\textbf{PANN} \\
\hspace{2mm} BERT & 0.02 ± 0.0 & 0.05 ± 0.01 & 0.01 ± 0.0 & 0.02 ± 0.01 & 0.0 ± 0.0 & 0.02 ± 0.01 \\
\hspace{2mm} BERT Large & 0.01 ± 0.0 & 0.04 ± 0.0 & 0.01 ± 0.0 & 0.03 ± 0.0 & 0.0 ± 0.0 & 0.01 ± 0.01 \\
\hspace{2mm} GPT-2 & 0.02 ± 0.01 & 0.04 ± 0.01 & 0.01 ± 0.0 & 0.02 ± 0.0 & 0.01 ± 0.01 & 0.03 ± 0.0 \\
\hspace{2mm} GPT-2 XL & 0.02 ± 0.0 & 0.06 ± 0.0 & 0.01 ± 0.0 & 0.03 ± 0.01 & 0.0 ± 0.0 & 0.01 ± 0.01 \\
\hspace{2mm} T5 & 0.01 ± 0.0 & 0.04 ± 0.01 & 0.01 ± 0.0 & 0.03 ± 0.0 & 0.01 ± 0.01 & 0.03 ± 0.01 \\
\hspace{2mm} T5 Large & 0.02 ± 0.0 & 0.05 ± 0.0 & 0.0 ± 0.0 & 0.02 ± 0.0 & 0.01 ± 0.01 & 0.02 ± 0.0 \\
\hspace{2mm} LLaMA-7B & 0.02 ± 0.0 & 0.06 ± 0.01 & 0.01 ± 0.0 & 0.03 ± 0.01 & 0.0 ± 0.0 & 0.03 ± 0.01 \\
\hspace{2mm} GloVe CC-42B & 0.01 ± 0.0 & 0.04 ± 0.01 & 0.0 ± 0.0 & 0.01 ± 0.0 & 0.0 ± 0.0 & 0.01 ± 0.01 \\
\hspace{2mm} GloVe CC-840B & 0.02 ± 0.01 & 0.05 ± 0.01 & 0.0 ± 0.0 & 0.01 ± 0.0 & 0.01 ± 0.01 & 0.01 ± 0.01 \\
\hspace{2mm} word2vec GNews-300 & 0.01 ± 0.0 & 0.03 ± 0.0 & 0.01 ± 0.0 & 0.03 ± 0.01 & 0.01 ± 0.01 & 0.01 ± 0.01 \\
\bottomrule
\end{tabular}
\caption{Numeric values for accuracy@1 (A@1) and accuracy@3 (A@3) for our main sound probing experiments. We also show standard error of the mean across 5 runs. Random init refers to a probe trained over randomly initialized language/audio models.}
\label{tab:full}
\end{table*}
\begin{figure*}[h]
\centering
\begin{subfigure}{0.45\textwidth}
    \centering
    \includegraphics[scale=0.17]{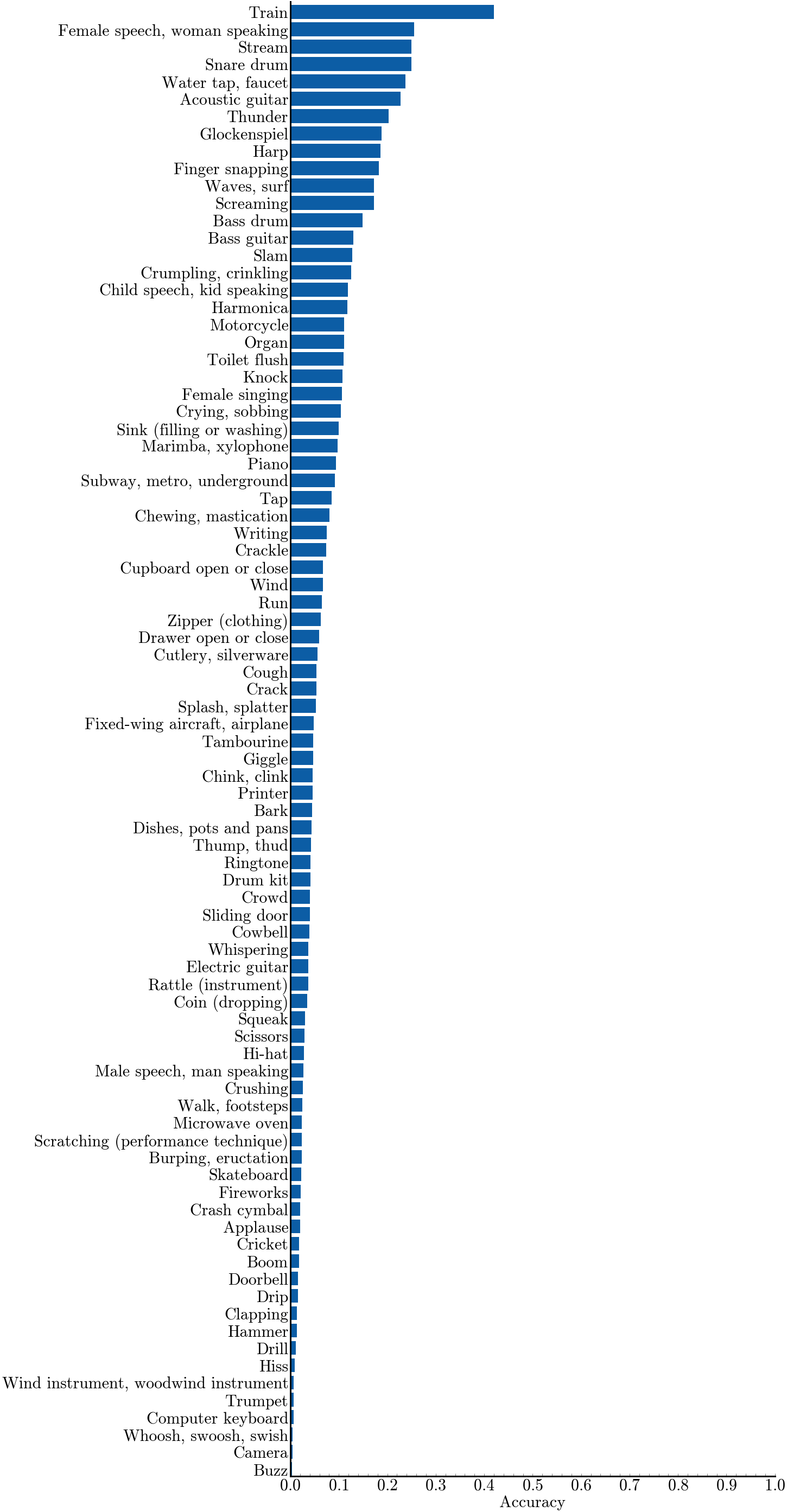}
    \caption{AudioMAE}
    \label{fig:first}
\end{subfigure}
\begin{subfigure}{0.45\textwidth}
    \centering
    \includegraphics[ scale=0.17]{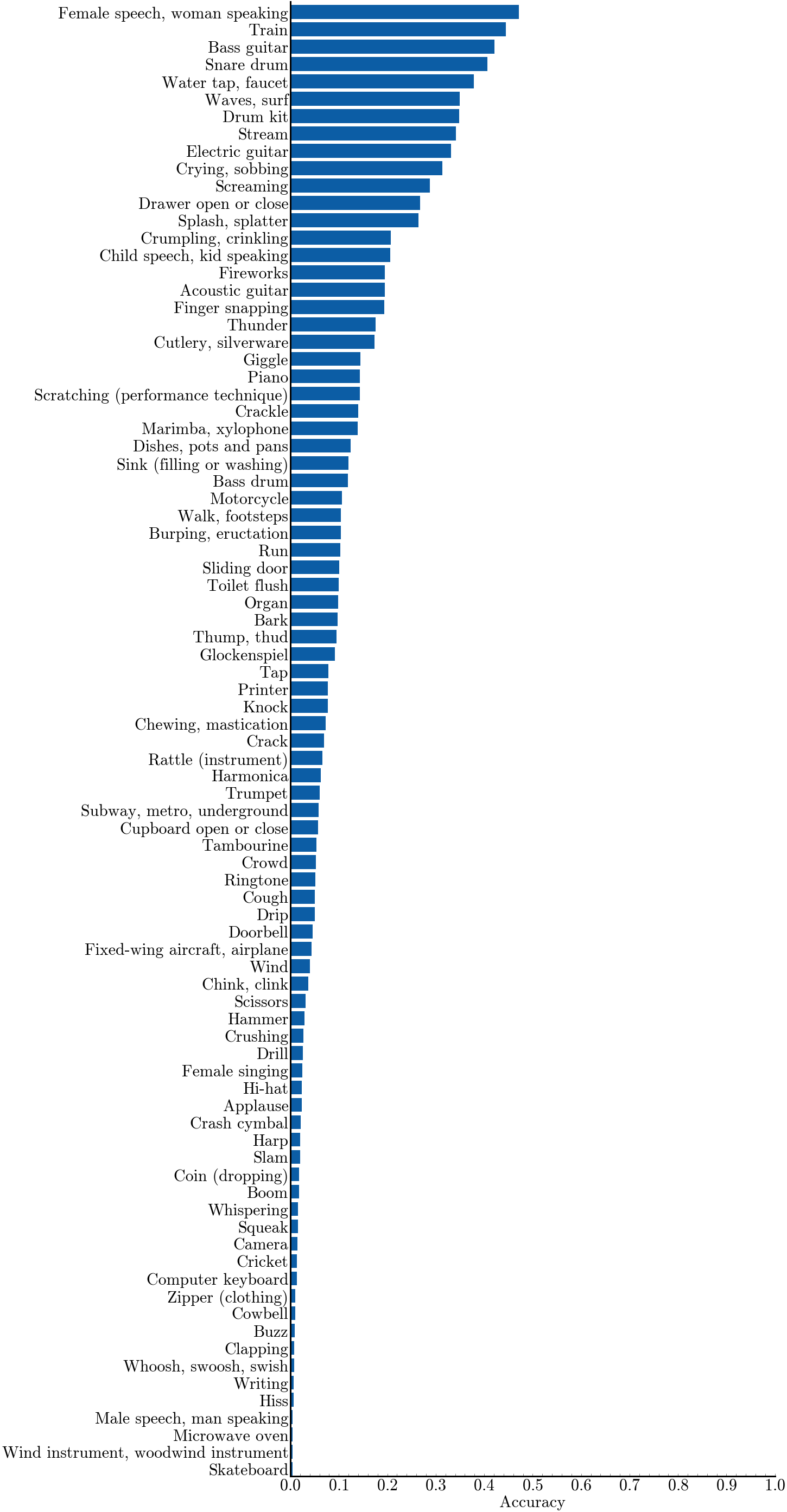}
    \caption{AudioMAE-FT}
    \label{fig:second}
\end{subfigure}
\end{figure*}
\begin{figure*}[t]
\ContinuedFloat 
\centering
\begin{subfigure}{0.45\textwidth}
    \centering
    \includegraphics[ scale=0.17]{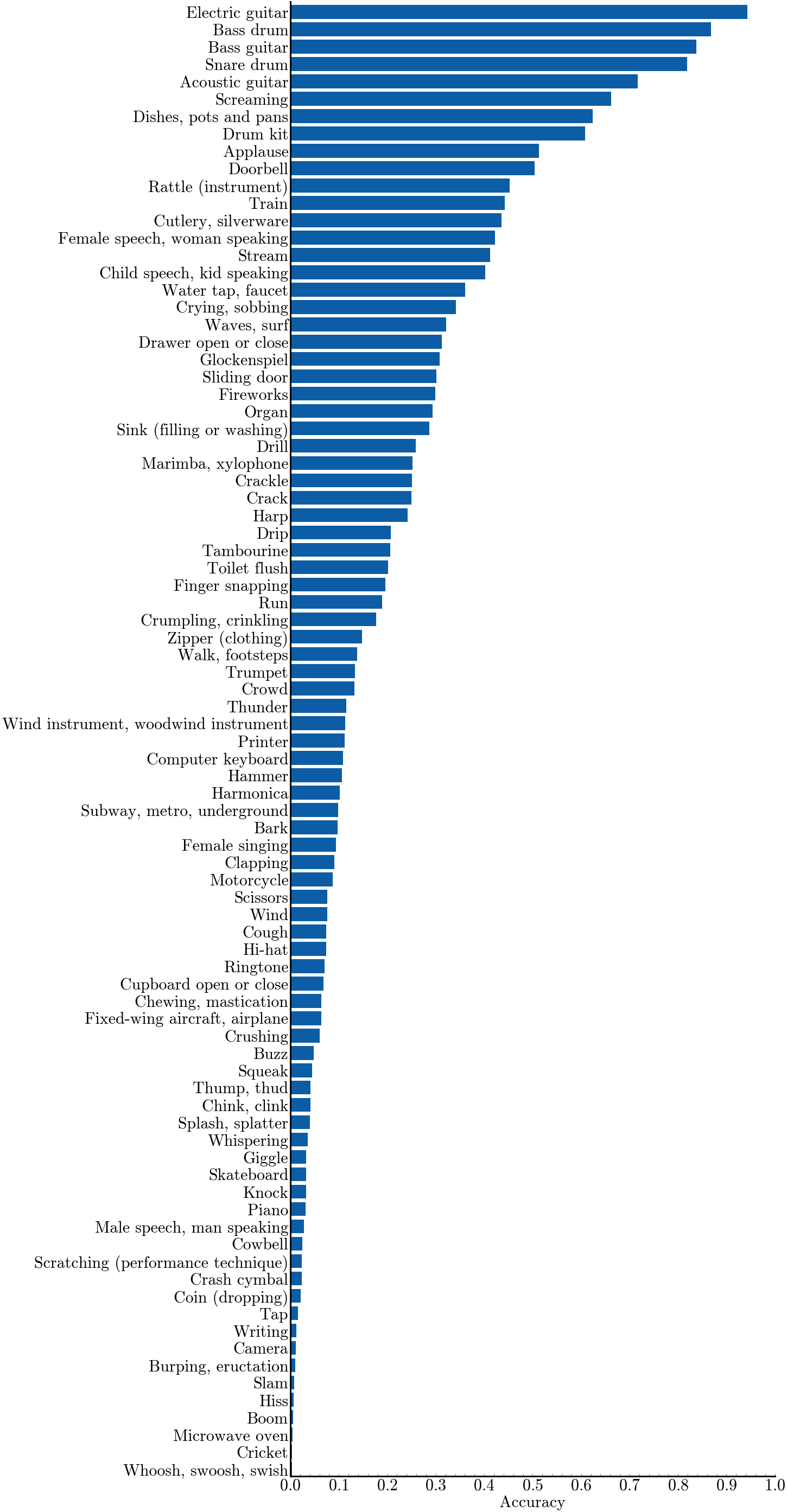}
    \caption{PaSST}
    \label{fig:third}
\end{subfigure}
\begin{subfigure}{0.45\textwidth}
    \centering
    \includegraphics[ scale=0.17]{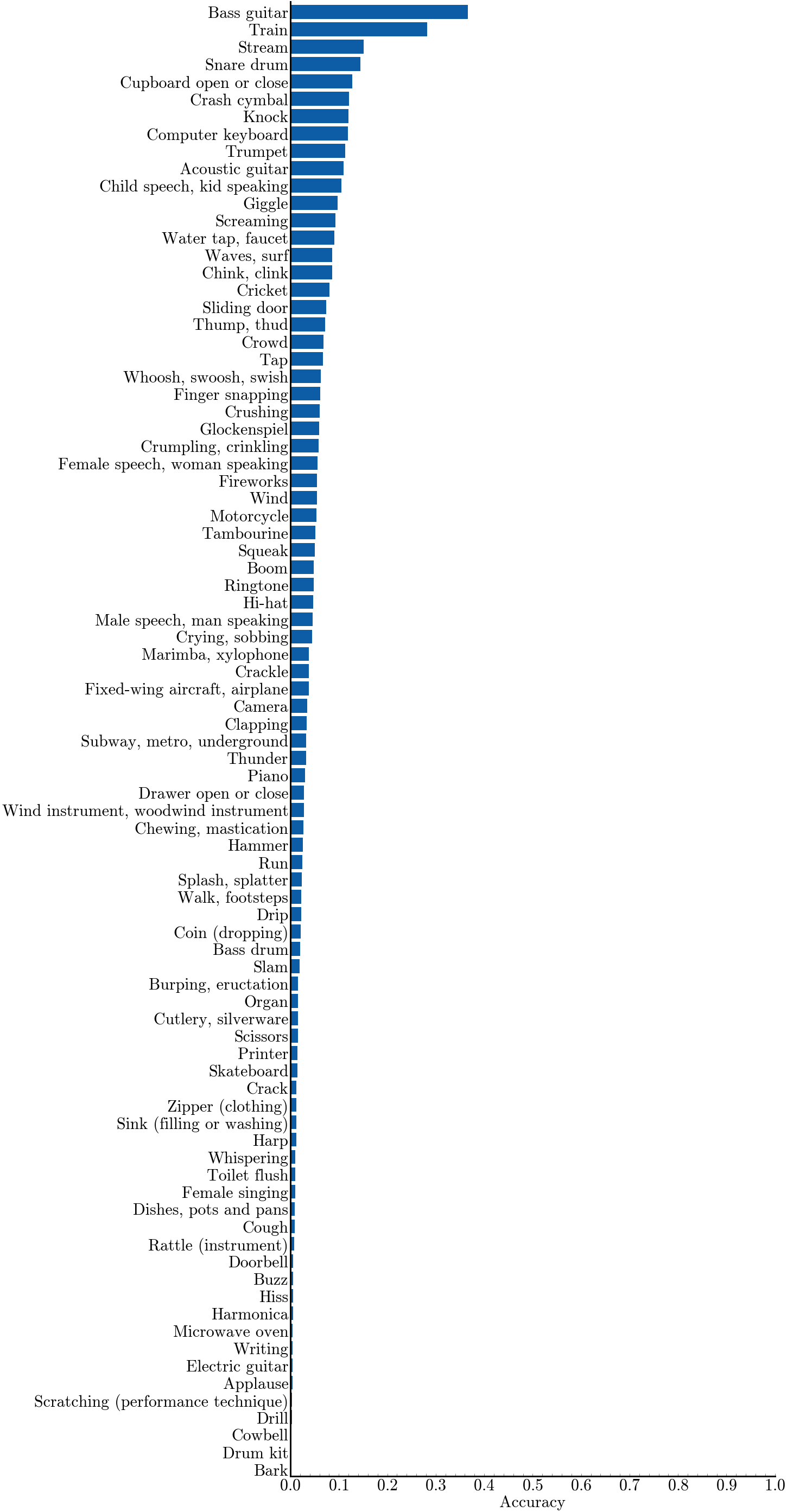}
    \caption{PANN}
    \label{fig:fourth}
\end{subfigure}
\vspace{-2mm}
\caption{All class accuracies for different audio models. Each class's accuracy is averaged across 5 train/test sets and different language models.}
    \label{fig:no_cut}
\end{figure*}

\end{document}